\documentclass[lettersize,journal]{IEEEtran}

\IEEEoverridecommandlockouts                             

\usepackage{amsmath}
\usepackage{amssymb}
  
\usepackage{amsthm}
\usepackage{graphicx}
\usepackage[Symbol]{upgreek}
\usepackage{epstopdf}
\usepackage{subcaption}
\usepackage{enumerate}
\usepackage{paralist}
\usepackage{float}
\usepackage{latexsym}
\usepackage{multicol}
\usepackage{multirow}
\usepackage{lipsum}
\usepackage{cite}
\usepackage{makecell}
\usepackage{breqn}
\usepackage{gensymb}
\usepackage{verbatim}
\usepackage{color}
\usepackage{xcolor}
\usepackage{bm}
\usepackage[font=small,labelfont=bf]{caption}
\usepackage{courier}
\usepackage{tikz}
\usetikzlibrary{calc}
\usepackage{arydshln}
\usepackage{nicefrac}
\usepackage{fancyhdr}

\newcommand{\vct}{\mathbf}
\newcommand{\vect}[1]{\boldsymbol{#1}}

\fancypagestyle{firstpage}
{
    
    \fancyhead[C]{\colorbox{yellow}{\rlap{This paper is a pre-print version under review}\hspace{\linewidth}\hspace{-2\fboxsep}}}    
}

\begin{document}

\title{Bimanual crop manipulation for human-inspired robotic harvesting}

\author{Sotiris Stavridis$^*$, Dimitrios Papageorgiou, Leonidas Droukas and Zoe Doulgeri 
\thanks{The research leading to these results has received funding from the European Community’s Framework Programme Horizon 2020 under grant agreement No 871704, project BACCHUS.} 
\thanks{Authors are with the Automation \& Robotics Lab, School of Electrical \& Computer Engineering, Aristotle University of Thessaloniki, Greece. Emails: {\tt\small \{sotistav@ece.auth.gr$^*$, dimpapag@ece.auth.gr, ldroukas@ece.auth.gr, doulgeri@ece.auth.gr\}}}
\thanks{$^*$Corresponding Author}
}

\maketitle

\thispagestyle{firstpage}


\begin{abstract}

Most existing robotic harvesters utilize a unimanual approach; a single arm grasps the crop and detaches it, either via a  detachment movement, or by cutting its stem with a specially designed gripper/cutter end-effector.
However, such unimanual solutions cannot be applied for sensitive crops and cluttered environments like grapes and a vineyard where obstacles may occlude the stem and leave no space for the cutter's placement.  In such cases, the solution would require a bimanual robot in order to visually unveil the stem and manipulate the grasped crop to create cutting affordances which is similar to the practice used by  humans.
In this work, a dual-arm coordinated motion control methodology for reaching a stem pre-cut state is proposed. The camera equipped arm with the cutter is reaching the stem, unveiling it as much as possible, while the second arm is moving the grasped crop towards the surrounding free-space to facilitate its stem cutting. Lab experimentation on a mock-up vine setup with a plastic grape cluster evaluates the proposed methodology, involving two UR5e robotic arms and a RealSense D415 camera.

\end{abstract}


\section{Introduction} \label{sec:Intro}

In the last decade, development of robotic technologies and their application in agricultural tasks are becoming a growing topic of interest \cite{Marinoudi}. The rising of food supply demands, climate change, land degradation and arable land limitations have all pushed towards agricultural productivity growth becoming an important priority. Incorporating advanced, automated, agricultural technologies is bound to greatly benefit productivity and in turn economic development \cite{Eberhardt}, while at the same time facilitating and improving the difficult working conditions of farmers and agricultural workers \cite{Fathallah}. In this context, an increasing amount of research work has been noticed during the last years \cite{Fountas,Bac}. Various agricultural tasks are considered as potential applications of robotic solutions developed in literature e.g. yield estimation, phenotyping, pruning, as well as crop harvesting. Earlier solutions in crop harvesting involved the bulk concept (i.e. tree trunk/branch shaking), however the selective concept has been mostly adopted lately, since it entails considerably less danger of harming the crops. In selective harvesting, the robotic system firstly detects a specific harvest target and then harvests it. 

\begin{figure}[h!]
	\centering
	\includegraphics[scale= .3]{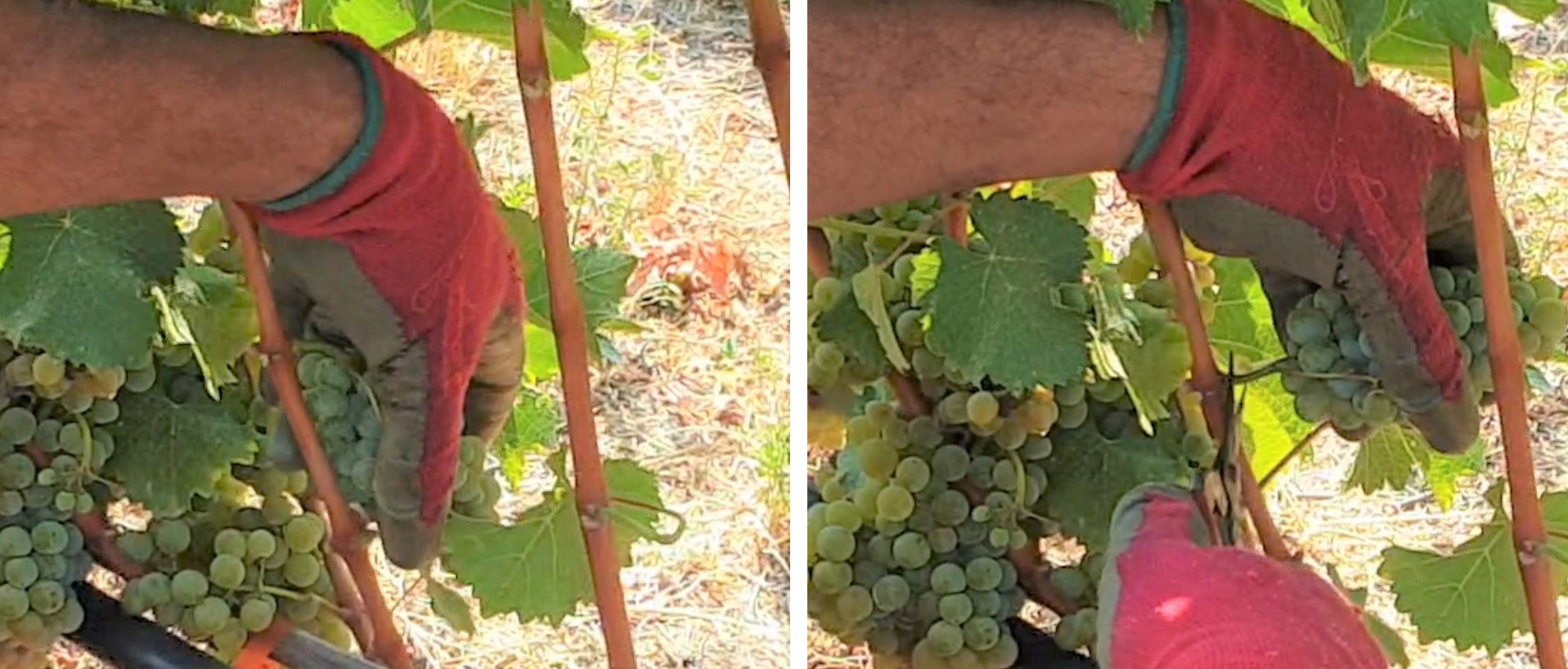}
	\caption{Bimanual human harvesting}
	\label{fig:human_harv}
\end{figure}

A variety of integrated robotic solutions have been proposed towards this end \cite{Ramin}, with the majority of them utilizing one robotic manipulator (unimanual setup), responsible for monitoring the target crop via a mounted in-hand camera and harvesting it, i.e. grasping it and detaching it. In \cite{Onishi}, \cite{Silwal} a detaching movement performed by the arm's gripper e.g. pulling and twisting the grasped crop in order to cause the stem's breaking and crop's release. However, detaching movements are only applicable on rigid crops (e.g. apples) that can withstand such treatment without being damaged. Alternatively, various approaches utilize the grasped crops stem cutting and mainly involve: (a) a cutting tool embodied in the gripper that cuts the stem after the crop is grasped \cite{sHayashi}, \cite{VanHenten}; (b) a specially designed cutter end-effector, responsible for simultaneously cutting and holding the crop by its stem \cite{Feng}, \cite{Hayashi2}. In these works, harvesting of eggplants, cucumbers, tomatoes and strawberries is performed. Such unimanual solution may work after certain crop modification e.g. pruning or cultivation/farming methodologies specifically designed to be more "friendly" towards automated/robotic harvesting such as the cucumber high-wire system \cite{VanHenten} and the apple V-trellis system \cite{Silwal}. However, in general, such unimanual solutions cannot be applied for sensitive and cluttered environments like a vineyard where poles, wires, flexible branches and leaves and even the crop itself partially occlude the stem, which can be further wrapped partially around a branch leaving no space for the cutter placement. In such cases the solution would require a bimanual robot in order to visually unveil the stem and manipulate the grasped crop to create cutting affordances. Notice that grape harvesting is performed bimanually by humans. In fact, a very common approach that humans apply during grape harvesting, is to grasp with one arm the crop, moving their head and body to be able to see the stem and manipulate the crop, if needed, in order to provide sufficient free room around the stem for the cutting tool held by their other arm (Fig. \ref{fig:human_harv}).  

There are few works in literature, proposing a dual-arm/bimanual solution \cite{Zhao,Ling}. In \cite{Zhao}, a
dual-arm tomato harvesting robot is designed, with two 3 degrees of freedom (DOF) SCARA manipulators (one prismatic and two revolute joints) and two different end-effectors (vacuum suction gripper and saw cutting tool). In \cite{Ling}, the same robotic solution is further developed, introducing autonomous ripe tomato detection in contrast to \cite{Zhao} where human collaboration was required, evaluated in a simple environment with several potted tomatoes placed in rows. However, the limited manipulability of the arms due to few DOF and the very specific gripper's design considering only rigid, spherical crops would render the above solution's application very difficult in more challenging crops/environments, such as grapes/vineyards.

In this work, we consider a bimanual robot with one arm equipped with a crop specific gripper (grasping arm) and the other with a stem cutting tool and a camera mounted on its end-effector (camera arm). Our focus is on the core bimanual task of approaching while unveiling the stem by the camera arm, while manipulating the grasped crop with the other arm for generating cutting affordances. Reaching to grasp the crop and reaching the final cut pose to cut the stem from a precut pose are mainly unimanual operations, which can be addressed by methods already proposed in the literature, like grasp planning \cite{Sahbani} and visual-servoing \cite{Ringdahl}.


\section{Problem Description} \label{sec:prob_descr} 

We consider a robot with two robotic arms: (a) the grasping arm of $N_g \in \mathbb{N}$ degrees of freedom (DOF) with $\vct{q}_g \in \mathbb{R}^{N_g}$ its joints vector, responsible for the stable grasping and manipulation of target crop; (b) the camera arm of $N_c \in \mathbb{N}$ DOF and joint vector $\vct{q}_c \in \mathbb{R}^{N_c}$ that monitors the whole scene via a mounted, in-hand, RGB-D camera equipped with a cutting tool. As an initial state for the robot we consider that the grasping arm has achieved grasping of the crop by its gripper and that a point-cloud of the overall scene is provided by the camera arm. We assume that an appropriate vision algorithm/methodology designed specifically for recognizing the stem is available. Notice that this assumption does not exclude the case of a stem being partially occluded or initially not recognized. 

Let $\vct{T}_c \in SE(3)$ be the homogeneous transformation expressing the generalized pose of the camera's frame $\{C\}$ (which is in our case placed at the camera arm's tip) with respect to the world inertial frame $\{0\}$; it involves its position $\vct{p}_c(\vct{q}_c) \in \mathbb{R}^3$ and its orientation $\vct{R}_c(\vct{q}_c) \triangleq [\vct{x}_c \; \vct{y}_c \; \vct{z}_c ] \in SO(3)$, that are both known via the camera arm's kinematics. Additionally, let $\vct{T}_g \in SE(3)$ be the homogeneous transformation expressing the grasping arm's tip/grasped crop frame $\{G\}$ with respect to $\{0\}$, involving position $\vct{p}_g(\vct{q}_g) \in \mathbb{R}^3$ and orientation $\vct{R}_g(\vct{q}_g) \triangleq [\vct{x}_g \; \vct{y}_g \; \vct{z}_g ] \in SO(3)$, known via the grasping arm's kinematics.

The core bimanual task of approaching and unveiling the stem with the camera arm and manipulating the grasped crop with the grasping arm, which is addressed in this work, is a bimanual asymmetric task where each arm's controller has a different objective but their motion needs appropriate planning and coordination. In particular, the camera arm's controller should operate in a clutter dynamical scene in which changes are not only induced by dynamic changes in the environment (e.g wind gusts can move the surrounding leaves) but also by the grasping arm's manipulation of the crop. In the following two subsections each arm's control objectives are presented.

\subsection{Camera arm control objectives} 
\label{subsec:cut_prob}

The camera arm is responsible for reaching a predefined region of interest (ROI) surrounding the crop and its stem and adopt a pose in this region that allows the stem to be unveiled as much as possible thus getting sufficiently close to it into a pre-cutting position. We can therefore distinguish two objectives that should be simultaneously satisfied by this controller, reaching and centering, and unveiling.

We model the ROI by a sphere and we call the associated stem, object of interest (OOI); let $\vct{p}_r \in \mathbb{R}^3$ be the center of ROI in the inertia frame with $r \in \mathbb{R}_{>0}$ its radius. The reaching and centering objective aims at moving the camera arm's end-effector to any of the ROI points, while aligning the camera's $Z$ axis with the vector pointing to the ROI center. The ROI center can be placed on a point on the stem if it is initially partially recognized or on the grape otherwise. As the camera moves closer and visual feedback provides better estimates of the point-cloud of the grape and its stem it is updated. To prevent discontinuities in the arm's operation the center of ROI displacement is smoothed by a filter. The two objectives can be formulated as follows: 
\begin{enumerate}
    \item \label{objective:reaching} \textit{Reaching and centering}: 
    
    for reaching the following should hold true: $\|\vct{p}_r - \vct{p}_c\| \rightarrow\overline{r} \leq r$ for $t\rightarrow\infty$ and some scalar positive $\overline{r}$; regarding centering, the angle between $\vct{p}_r - \vct{p}_c$ and the camera's view direction $\vct{z}_c$ defined as $\theta(\vct{p}_c, \vct{p}_r) \triangleq \text{cos}^{-1}\left(\frac{\vct{z}_c^\intercal(\vct{p}_r - \vct{p}_c)}{\|\vct{p}_r - \vct{p}_c\|}\right)$ should satisfy $\theta(\vct{p}_c, \vct{p}_r)\rightarrow 0$ for $t\rightarrow\infty$.
    
    \item \label{objective:unveiling} \textit{Unveiling}: maximize the visible part of the OOI, i.e. maximize the perceived number of points of the point-cloud that belong to the stem. 
\end{enumerate}
 Notice that $\theta \rightarrow 0$ involved in the first objective maintains the ROI center at the center of the camera's field of view, thus securing the best ROI and OOI viewpoint. Moreover, notice that $\theta$ cannot be more than $\nicefrac{\pi}{2}$, since the ROI center $\vct{p}_r$ is within the camera's field of view, i.e. a pyramid. 
 
 Clearly, the reaching and centering objective is achieved by reaching any position in ROI which is associated with two specific orientation dof for centering.  The unveiling objective \eqref{objective:unveiling} selects the position in ROI that visually unveils the OOI as much as possible. Hence, all the above objectives correspond to 5 dof with the rotation around the z camera axis denoting a task redundancy of 1 dof. The latter can be specified for properly rotating the cutter wrt the stem.

\subsection{Grasping Arm control objective} 
\label{subsec:grasp_prob}

Regarding the grasping arm's task, the goal is to maximize the free room around the stem  to enable the placement of the cutter. This is achieved by manipulating the grasped grape so that the stem is stretched and moved towards a point in the free space that is most distant from any surrounding obstacle identified from the provided scene's point-cloud. Let this point's position be  $\vct{p}_{gd}$. We assume that  $\vct{f} \in \mathbb{R}^3$, the actual applied force vector at grasping arm's tip, is measurable. Notice that the grape is an object  hinged to the branch by its stem. We can estimate the stem basis from the point-cloud and let $\vct{n}_c \in \mathbb{R}^3$ be the direction from the stem basis to $\vct{p}_{gd}$. We can then formulate the control objective as a force position control objective along $\vct{n}_c$ and its orthogonal subspace respectively.  Hence if  $f_d \in \mathbb{R}$ is an appropriately selected force magnitude for stretching the stem sufficiently without damaging/tearing it, the grasping arm's control objectives can be mathematically expressed as follows: $\vct{n}_c \vct{n}_c^\intercal \vct{f} \rightarrow \vct{n}_c f_d$ and $\vct{p}_{g} \rightarrow (\vct{I}_3-\vct{n}_c \vct{n}_c^\intercal)\vct{p}_{gd}$.

Notice that we have decomposed the space of the grasping arm's end-effector position  based on the identified $\vct{p}_{gd}$ and thus both  the position and force control action space can create motion until the final target is reached. Alternatively the space should be decomposed appropriately given the current position of the grape and its stem. As compared to the alternative decomposition solution, the adopted approach is computationally lighter and shown to be effective as we will demonstrate in the experimental part.
  

\section{Proposed Control Methodology} \label{sec:contr}

We consider a velocity controlled bimanual robot. Hence our control objective is to design a reference velocity control signal $\vct{V}_r = [\vct{V}_c^\intercal \ \ \vct{V}_g^\intercal]^\intercal \in \mathbb{R}^{12}$ involving the arm end-effector velocities $\vct{V}_c, \ \vct{V}_g \in \mathbb{R}^6$ for the cutting and grasping arm respectively, which can then be mapped to the joint space via the robot's extended Jacobian matrix $\vct{J}(\vct{q}) = diag(\vct{J}_c(\vct{q}_c) , \vct{J}_g(\vct{q}_g)) \in \mathbb{R}^{12 \times (N_c + N_g)}$ involving the camera arm Jacobian $\vct{J}_c(\vct{q}_c) \in \mathbb{R}^{N_c \times 6}$ and the grasping arm Jacobian $\vct{J}_g(\vct{q}_g) \in \mathbb{R}^{N_g \times 6}$  as follows:
  
\begin{equation} \label{eq:controlled_system}
    \dot{\vct{q}} = \vct{J}^\dagger(\vct{q}) \vct{V}_r .
\end{equation}
where  $\vct{q} = [\vct{q}_c^\intercal \ \ \vct{q}_g^\intercal]^\intercal \in \mathbb{R}^{N_c + N_g}$ denotes the robot's joint position, $\vct{J}^\dagger(\vct{q})$ is a pseudo inverse of the robot's extended Jacobian matrix.

Designing $\vct{V}_c$ and $\vct{V}_g$ is heavily based upon the overall scene's point-cloud set and its processing in order to estimate the required quantities involved in each control objective as well as to identify the point-cloud of the stem and the surrounding obstacles. 

\subsection{Scene Point-cloud Process and   calculations of control related critical point positions} 
\label{subsec:scePC_proc}

\begin{figure}[t!]
    \begin{subfigure}[t]{.48\linewidth }
        \centering
        \includegraphics[width = 1.1\linewidth]{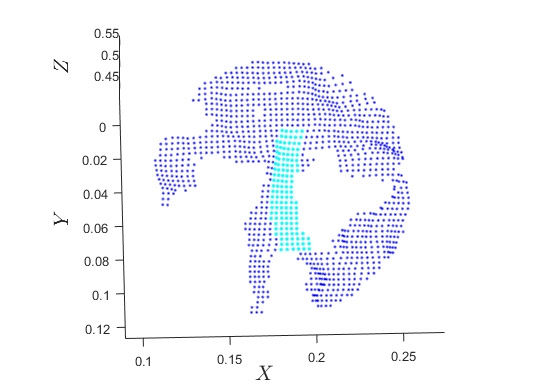}
        \caption{Scene's point-cloud $\mathcal{W}$. Obstacles subset $\mathcal{O}$: blue. Stem subset $\mathcal{S}$: cyan.}
        \label{fig:PC_proc_1}
    \end{subfigure}
    \hfill 
    \begin{subfigure}[t]{.48\linewidth }
        \centering
        \includegraphics[width = 1.1\linewidth]{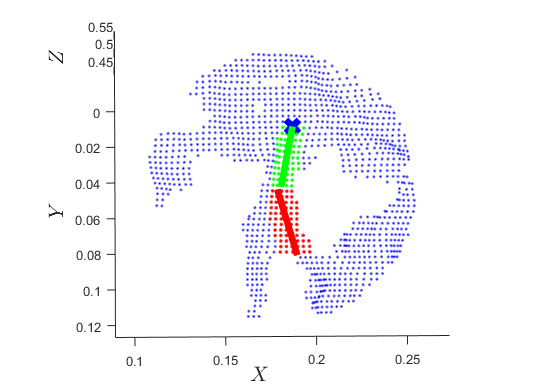}
        \caption{Stem's point-cloud clustering and line segment fitting: green and red. Stem's base: blue \textbf{\textcolor{blue}{x}}.}
        \label{fig:PC_proc_2}
    \end{subfigure}
    \vskip \baselineskip
    \begin{subfigure}[t]{.48\linewidth}
        \centering
        \includegraphics[width = 1.1\linewidth]{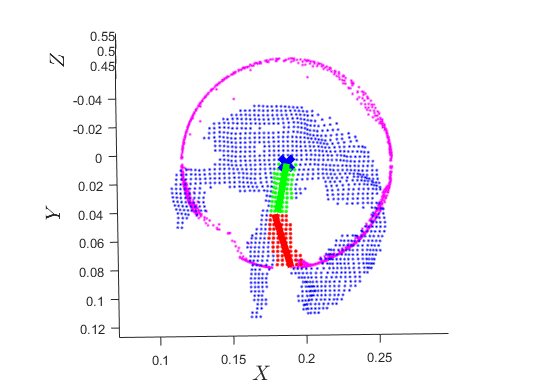}
        \caption{Obstacles projection upon sphere subset $\mathcal{O}_{pr}$: purple.}
        \label{fig:PC_proc_3}
    \end{subfigure}
    \hfill 
    \begin{subfigure}[t]{.48\linewidth}
        \centering
        \includegraphics[width = 1.1\linewidth]{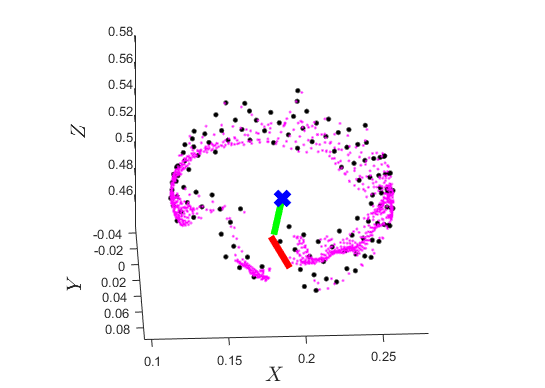}
        \caption{Sphere surface sampling ($500$ samples). Near-obstacles sampled points: black. $\mathcal{O}$ and $\mathcal{S}$ omitted for clarity.}
        \label{fig:PC_proc_4}
    \end{subfigure}
    \vskip \baselineskip
    \begin{subfigure}[t]{.48\linewidth}
        \centering
        \includegraphics[width = 1.1\linewidth]{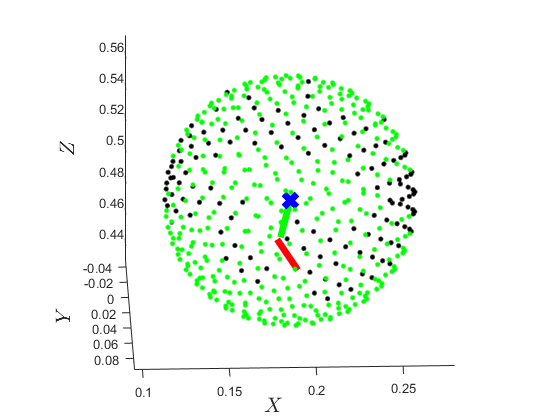}
        \caption{Free-space sphere points: green. Near-obstacles sampled points: black. $\mathcal{O}_{pr}$ omitted for clarity.}
        \label{fig:PC_proc_5}
    \end{subfigure}
    \hfill 
    \begin{subfigure}[t]{.48\linewidth}
        \centering
        \includegraphics[width = 1.1\linewidth]{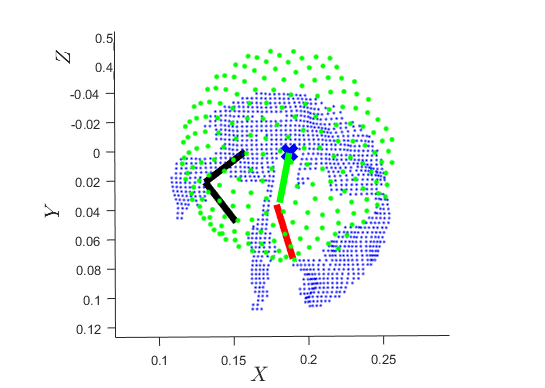}
        \caption{Free-space hemisphere subset $\mathcal{F}$: green. Obstacles $\mathcal{O}$: blue. Plane fitted into $\mathcal{O}$: black lines. Near-obstacles sampled points omitted for clarity.}
        \label{fig:PC_proc_7}
    \end{subfigure}
    \caption{Point-cloud processing example from lab experimentation with a Real-Sense camera and a mock-up crop. Figures shown from camera's viewpoint with $Z$ axis corresponding to the camera's view direction.}
    \label{fig:PC_proc}
\end{figure}

Let $\mathcal{W}: \{ \vct{p}_{w_i} \in \mathbb{R}^3 \ , \ i=1,...,n_W \}$ be the set of $n_W \in \mathbb{N}$ points of the point-cloud captured by the camera  with $\vct{p}_{w_i}$ being their position vector  expressed in the world frame (Fig. \ref{fig:PC_proc_1}). Let subset $\mathcal{S} \subseteq \mathcal{W}: \{ \vct{p}_{s_j} \in \mathbb{R}^3 \ , \ j=1,...,n_S \}$ denote the detected stem's point-cloud of $n_S \in \mathbb{N}$ points of position $\vct{p}_{s_j}$ (Fig. \ref{fig:PC_proc_1}, cyan points). To identify the stem's base $\vct{p}_{sb} \in \mathbb{R}^3$, we model the stem by connected linear segments to account for their bending capability. To this end we initially partition $\mathcal{S}$ into separate, consecutive point-cloud clusters, utilizing any of the existing clustering methodologies, e.g. the k-means clustering algorithm. Since crop stems are generally short (e.g. $5-7$ cm), two clusters are considered to be enough to adequately model the stem's flexibility. We then identify which cluster is further and closer to the grasping arm, comparing the distance between the gripper and each cluster mean value and denote them as "top" $\mathcal{T}: \{ \vct{p}_{cl_a} \in \mathbb{R}^3 \ , \ a=1,...,n_T \}$ and "bottom" $\mathcal{B}: \{ \vct{p}_{cl_b} \in \mathbb{R}^3 \ , \ b=1,...,n_B \}$ cluster respectively where $\vct{p}_{cl_a}, \ \vct{p}_{cl_b}$ and $n_T,n_B$ are their respective points' position vectors and number, with $\mathcal{T},\mathcal{B} \subseteq \mathcal{S}$ and $\mathcal{T} \neq \mathcal{B}$. Utilizing the Principal Component Analysis (PCA), a line segment is fitted into each cluster. Let  $\vct{n}_{st}, \ \vct{n}_{sb} \in \mathbb{R}^3$ be their direction vectors found by the corresponding major PCA eigenvectors. We can also find the line segments'  length $l_t, \ l_b \in \mathbb{R}_{>0}$ for the top and bottom line respectively, and deduce an estimate of the whole stem's length $l = l_t + l_b \in \mathbb{R}_{>0}$.  

\begin{table}[b!]
    \centering
        \begin{tabular}{ |l|l|  }                             \hline
            Point-cloud Subsets & Involved Points          \\ \hline 
            $\text{whole scene}: \mathcal{W}$                                         
            & $\vct{p}_{w_i}  \in \mathbb{R}^3 \ , \ i=1,...,n_W$ \\[1ex]
            $\text{stem}:        \mathcal{S} \ , \ \mathcal{S} \subseteq \mathcal{W}$ 
            & $\vct{p}_{s_j}  \in \mathbb{R}^3 \ , \ j=1,...,n_S$ \\[1ex]
            $\text{stem top}:    \mathcal{T} \ , \ \mathcal{T} \subseteq \mathcal{S}$ 
            & $\vct{p}_{cl_a} \in \mathbb{R}^3 \ , \ a=1,...,n_T$ \\[1ex]
            $\text{stem bottom}: \mathcal{B} \neq \mathcal{T} \ , \ \mathcal{B} \subseteq \mathcal{S}$ 
            & $\vct{p}_{cl_b} \in \mathbb{R}^3 \ , \ b=1,...,n_B$ \\[1ex]
            $\text{obstacles}:   \makecell[l]{\mathcal{O} \triangleq (\mathcal{W} - \mathcal{S}) \cap \\ S_{ph}(\vct{p}_{sb},r_o)}$ 
            & $\vct{p}_{o_k}  \in \mathbb{R}^3 \ , \ k=1,...,n_O$ \\[1ex]
            $\text{proj. obstacles}: \makecell[l]{\mathcal{O}_{pr} \triangleq \textit{proj}(\mathcal{O}) \ \\ \text{upon} \ \text{S}(\vct{p}_{sb},l)}$     
            & $\textit{proj}_{\text{S}}(\vct{p}_{o_k})  \in \mathbb{R}^3 \ , \ k=1,...,n_O$ \\[1ex]
            $\text{free-space}:  \mathcal{F}$  
            & $\vct{p}_{f_m}  \in \mathbb{R}^3 \ , \ m=1,...,n_F$ \\[1ex] \hline
        \end{tabular}
    \caption{Point-cloud Classification.}
    \label{table:PCs}
\end{table}

The stem's base $\vct{p}_{sb}$ is then calculated as follows (Fig. \ref{fig:PC_proc_2}): 
\begin{align} \label{stem_basend}
    \vct{p}_{sb} &= \text{mean}(\mathcal{T}) - (l_t/2)\vct{n}_{st} 
\end{align}
where $\text{mean}(X)$ is the mean value of $X$ point-cloud subset. For simplicity, we consider that target crop's stem is attached to a branch with limited mobility. Hence, the stem's base $\vct{p}_{sb}$ is assumed to be static and is identified only once at the start of the reaching/unveiling task. 

All scene points that do not belong to $\mathcal{S}$ can be considered as potential obstacles but we limit our search for obstacles in an area closely around the stem, i.e. a sphere $\text{S}(\vct{p}_{sb},r_o)$ with the stem's base $\vct{p}_{sb}$ as its center and a radius of $r_o \in \mathbb{R}$, selected freely to be adequately close to the stem. Thus, the obstacle point-cloud subset consists of $n_O \in \mathbb{N}$ points, that are not stem points and reside inside the above considered sphere, with vector $\vct{p}_{o_k}$ describing their position: $\mathcal{O}: \{ \vct{p}_{o_k} \in \mathbb{R}^3 \ , \ k=1,...,n_O \}$, with $\mathcal{O} \triangleq (\mathcal{W} - \mathcal{S}) \ \cap \ \text{S}(\vct{p}_{sb},r_o)$ (Fig. \ref{fig:PC_proc_1}, blue points).

Let subset $\mathcal{F}: \{ \vct{p}_{f_m} \in \mathbb{R}^3 \ , \ m=1,...,n_F \}$ denote the free-space surrounding the stem, containing $n_F \in \mathbb{N}$ points with position vector $\vct{p}_{f_m}$. In order to calculate $\mathcal{F}$, we define a sphere $\text{S}(\vct{p}_{sb},l)$ with the stem's base $\vct{p}_{sb}$ as its center and a radius equal to the calculated stem's length $l$. Subset $\mathcal{O}$ points are then projected upon this sphere's surface with $\mathcal{O}_{pr}: \{ \textit{proj}_{\text{S}}(\vct{p}_{o_k}) \in \mathbb{R}^3 \ , \ k=1,...,n_O \}$ denoting the set of these projected points with $\textit{proj}_{\text{S}}(\vct{p}_{o_k})$ their position vector (Fig. \ref{fig:PC_proc_3} - \ref{fig:PC_proc_4}, purple points) given by: $\textit{proj}_{\text{S}}(\vct{p}_{o_k}) = \frac{\vct{p}_{o_k} - \vct{p}_{sb}}{\|\vct{p}_{o_k} - \vct{p}_{sb}\|}l + \vct{p}_{sb}$. Sphere $\text{S}(\vct{p}_{sb},l)$ is uniformly sampled utilizing the Fibonacci lattice methodology, which is a fast and near optimal way of sampling a sphere. Furthermore, in \cite{fibonaci} a method to find the closest Fibonacci point for a given point upon the sphere is introduced, allowing the match of the sphere projected obstacle points $\mathcal{O}_{pr}$ to corresponding Fibonacci points without resorting to an exhaustive search that would significantly increase the calculation cost. Hence assuming adequate sampling, the sampled sphere points nearest to $\mathcal{O}_{pr}$ points are found and then removed from the sphere's surface, leaving the remaining sphere surface points which constitute the free-space surrounding the crop's stem (Fig. \ref{fig:PC_proc_5}). 
Since the camera cannot see behind obstacles (e.g. leaves, target crop), we apply PCA on subset $\mathcal{O}$ to calculate the corresponding two major eigenvectors i.e. the two direction vectors where most $\mathcal{O}$ data reside; this two vectors define a representative plane fitted into $\mathcal{O}$ (Fig. \ref{fig:PC_proc_7}, black lines). Any free-space point upon the sphere's surface which is behind this plane, is considered invalid. Hence, the previously defined sphere becomes a hemisphere, ensuring that the determined free-space $\mathcal{F}$ is deduced so that it always resides within the camera's clearly visible field of view (Fig. \ref{fig:PC_proc_7}). 
From $\mathcal{F}$, free-space point $\vct{p}_{gd}$, which is tied to the grasping arm's control objective (Subsection \ref{subsec:grasp_prob}), is calculated by solving the following optimization problem:
\begin{equation} \label{eq:attr_point}
    \vct{p}_{gd} = \operatorname*{argmax}_{\vct{p}_{f_m}} \Big\{ \min_{\forall k} \big\{ \|\vct{p}_{f_m} - \textit{proj}_{\text{S}}(\vct{p}_{o_k})\| \big\} \Big\},
\end{equation}
utilizing the projected obstacle points $\textit{proj}_{\text{S}}(\vct{p}_{o_k})$ upon the sphere $\text{S}(\vct{p}_{sb},l)$. Given a relatively low number of free-space points by appropriate selection of the sphere's sampling rate, \eqref{eq:attr_point} can be solved with brute force in a short amount of time. Table \ref{table:PCs} summarizes the classified/calculated point-cloud subsets.

\subsection{Camera arm Control - Reaching \& Unveiling Tasks} 
\label{subsec:cut_contr}

The camera arm end-effector's reference velocity is calculated by the superposition of two reference velocity control terms, the ROI reaching with centering control signal $\vct{V}_{cr} \in \mathbb{R}^6$ and the  OOI unveiling control signal, $\vct{V}_{cu} \in \mathbb{R}^6$:
\begin{equation} \label{eq:superimposition}
    \vct{V}_c = \vct{V}_{cr} + \vct{V}_{cu},
\end{equation}
When there are no stem points detected ($n_S = 0$) $\vct{V}_{cu} = \vct{0}_{6 \times 1}$  by design and  $\vct{V}_{cr}$ is acting alone. Such a case may occur during the start of the motion, when the camera is far from the ROI and the OOI starts being visible only when it approaches or enters the ROI. Figure \ref{fig:cutt_contr} depicts a typical  initial and final state of the camera's motion under the superimposed velocities.
\newline

\subsubsection{Reaching reference velocity} 
\label{subsec:rr_control_signal}

Mathematically, reaching is fulfilled when the camera position $\vct{p}_c$ converges to the manifold $\Omega \triangleq \{ \vct{p}_c \in \mathbb{R}^3: \text{f}(\vct{p}_r - \vct{p}_c) \leq 0 \}$, with $\text{f}(\vct{x}) \triangleq \vct{x}^\intercal \vct{x} - r^2$ for any $\vct x \in \mathbb{R}^3$. Centering the camera's field of view with respect to ROI center regards the orientation of the camera and is achieved when the camera's $Z$ axis points to the center of the ROI (Fig. \ref{fig:cutt_contr}). To achieve the reaching with centering objective we propose the following control signal, which draws its inspiration from the region reaching approach proposed in \cite{Cheah2005} for the translational part:
\begin{equation} \label{eq:control_signal_rr_world}
    \vct{V}_{cr} \triangleq 
    \begin{bmatrix}
       k_{cp} \text{max}(0,\; \text{f}(\vct{e}_{pc}))\vct{e}_{pc} \\
       k_{co} \theta \vct{k}
    \end{bmatrix},
\end{equation}
where $k_{cp}, k_{co} \in\mathbb{R}_{>0}$ are constant positive gains for translation and orientation respectively and $\vct{e}_{pc} = \vct{p}_r - \vct{p}_c$; $\theta\in[0,\nicefrac{\pi}{2}]$ and $\vct k$ are the angle and axis of the minimum rotation between $\vct{e}_{pc}$ and $\vct{z}_c$, which can be calculated by the following expressions:
\begin{equation} \label{eq:ktheta_world}
    \theta = \text{cos}^{-1} \left( \frac{\vct{z}_c^\intercal \vct{e}_{pc}}{\|\vct{e}_{pc}\|}\right), \;
    \vct{k} = \frac{\vct{S}( \vct{z}_c ) \vct{e}_{pc}}{\|\vct{S}( \vct{z}_c )\vct{e}_{pc}\|}
\end{equation}
with $\vct{S}(\vct{z})$ the skew symmetric matrix of the corresponding vector $\vct{z}$. Notice that $\vct{V}_{cr}$ is continuous and its value is zero if and only if $\text{f}(\vct{e}_{pc}) \leq 0$ and $\theta = 0$. When this signal is acting alone it is easy to prove asymptotic convergence to the manifold $\text{f}(\vct{e}_{pc}) \leq 0$ and $\theta = 0$. 

\subsubsection{Unveiling Reference velocity} 
\label{subsec:occlusion_control_signal}

\begin{figure}[!htbp]
    \centering
    \includegraphics[scale = 0.45]{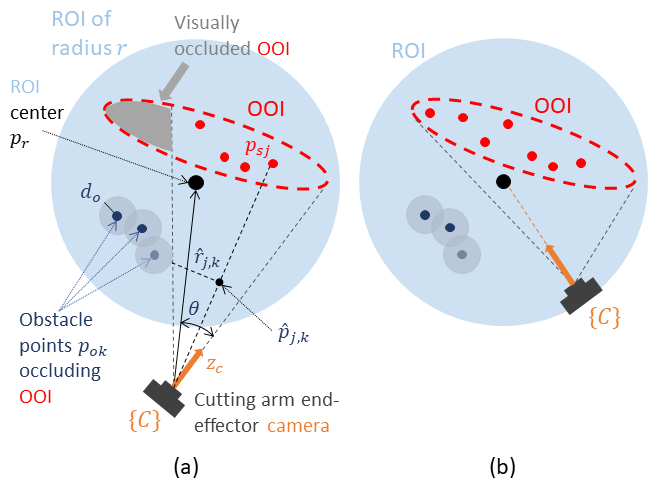}
    \caption{Camera arm control methodology. (a) Initial state. (b) Desired state.} 
    \label{fig:cutt_contr}
\end{figure}

Regarding unveiling, we consider obstacles not just as points but as spheres with radius $d_o \in \mathbb{R}$ selected such that the empty space between neighboring points of the obstacle point-cloud subset $\mathcal{O}$ is covered (Fig. \ref{fig:cutt_contr}). Notice the ray from the camera  to  a visible point of the OOI, $\vct{p}_{s_j}$, shown in Fig. \ref{fig:cutt_contr}. Let $\hat{\vct{p}}_{j,k}$ be a point in this ray that has a minimum distance from an obstacle point $\vct{p}_{o_k}$. Let $\hat{r}_{j,k} \in \mathbb{R}_{\geq 0}$ be the distance from the obstacle surface i.e.  $\hat{r}_{j,k} \triangleq \|\hat{\vct{p}}_{j,k} - \vct{p}_{o_k}\| - d_o$. The basic concept of the unveiling reference velocity is targeting at rotating the camera around $\vct{p}_{s_j}$ in order to guarantee that $\hat{r}_{j,k}$ will increase and $\hat{r}_{j,k} > d_o$ so that the unveiled region around $\vct{p}_{s_j}$ will also increase.

Let $\vct{p}_{s_j} \ , \ j=1,...,n_S$ be the currently visible OOI points and $\vct{p}_{o_k} \ , \ k=1,...,n_O$ obstacles.
The nearest point of the $j$-th ray from the $k$-th obstacle is calculated as follows:
 \begin{equation}\label{eq:nearest}
    \hat{\vct{p}}_{j,k} \triangleq \begin{cases}
        \vct{p}_c,\; &\text{if} \; \Big( \frac{\vct{\bar{p}}_{cs_j}^\intercal}{\|\vct{\bar{p}}_{cs_j}\|} \Big) \vct{\bar{p}}_{co_k} \leq 0 \\
        
        \vct{p}_c + \frac{\vct{\bar{p}}_{cs_j}\vct{\bar{p}}_{cs_j}^\intercal}{\vct{\bar{p}}_{cs_j}^\intercal\vct{\bar{p}}_{cs_j}} \vct{\bar{p}}_{co_k}, \; &\text{if} \; \Big( \frac{\vct{\bar{p}}_{cs_j}^\intercal}{\|\vct{\bar{p}}_{cs_j}\|} \Big) \vct{\bar{p}}_{co_k} \in(0, \|\vct{\bar{p}}_{cs_j}\|) \\
        
        \vct{p}_{s_j}, \; &\text{if} \; \Big( \frac{\vct{\bar{p}}_{cs_j}^\intercal}{\|\vct{\bar{p}}_{cs_j}\|} \Big) \vct{\bar{p}}_{co_k} \geq \|\vct{\bar{p}}_{cs_j}\| \\
    \end{cases}
\end{equation}
with $\vct{\bar{p}}_{cs_j} = \vct{p}_{s_j} - \vct{p}_{c}$ and $\vct{\bar{p}}_{co_k} = \vct{p}_{o_k} - \vct{p}_{c}$.

Drawing our inspiration from our previous work \cite{Kastritsi2019}, we propose the utilization of a barrier artificial potential field around each obstacle, which induces a virtual repulsive velocity $\vct{u}_{j,k}$, acting at $\hat{\vct{p}}_{j,k}$. 
The proposed barrier artificial potential function 
is designed to induce a repulsive velocity only within a predefined distance $d_a \in \mathbb{R}_{>0}$ from the obstacle surface and is defined as follows:
\begin{equation}\label{eq:art_potential}
    V(\hat{r}_{j,k})\triangleq 
    \begin{cases}
    \frac{1}{2}\ln^2\left(\frac{d_a^2}{d_a^2-(d_a-\hat{r}_{j,k})^2}\right), \; &\text{if} \; \hat{r}_{j,k}< d_a \\
    0, \; & \text{otherwise} 
    \end{cases},
\end{equation}
Notice that the value of the artificial potential tends to infinity, when the surface of the obstacle is reached from outside in order to guarantee obstacle/occlusion avoidance.

The virtual commanded repulsive velocity is then given by:
\begin{equation}\label{eq:u_ij}
\begin{split}
    &\vct{u}_{j,k} \triangleq - \frac{\partial V(\hat{\vct{p}}_{j,k})}{\partial\hat{\vct{p}}_{j,k}} = \\
    &\begin{cases}
    \frac{2}{d_a^2-(d_a-\hat{r}_{j,k})^2}\ln\left(\frac{d_a^2}{d_a^2-(d_a-\hat{r}_{j,k})^2}\right)\vct{e}_{j,k}, \;  &\text{if} \; \hat{r}_{j,k}< d_a \\
    \vct{0}_{3 \times 1}, \; &\text{otherwise} 
    \end{cases}
    \end{split}
\end{equation}
with $\vct{e}_{j,k} \triangleq (d_a-\hat{r}_{j,k})\frac{\hat{\vct{p}}_{j,k} - \vct{p}_{o_k}}{\|\hat{\vct{p}}_{j,k} - \vct{p}_{o_k}\|} \in \mathbb{R}^3$. The virtual repulsive velocity \eqref{eq:u_ij} possesses the following properties:
\begin{itemize}
    \item $\|\vct{u}_{j,k}\| \neq 0$, if and only if $0 < \hat{r}_{j,k} < d_a$, and $\vct{u}_{j,k} \neq \vct{0}_{3 \times 1}$ if and only if $\hat{r}_{j,k} \geq d_a$, which means that the signal will not be affected by rays that are not within the range of influence of the $k$-th obstacle point, defined by $d_a$.
    \item $\|\vct{u}_{j,k}\| \rightarrow \infty$, when $\hat{r}_{j,k} \rightarrow 0$, i.e. when the $j$-th ray approaches the surface of the $k$-th obstacle. Notice that $\|\hat{\vct{p}}_{j,k} - \vct{p}_{o_k}\|$ cannot be less than $d_o$, as it reflects the accuracy of the RGB-D camera, by definition.
    \item $\vct{u}_{j,k}$ is continuous with respect to $\hat{\vct{p}}_{j,k}$.
    \item  $\vct{u}_{j,k}$ is, in general, not continuous in time, as $\vct{p}_{s_j}, \vct{p}_{o_k}$ and even $n_S, \ n_O$ depend on the point-cloud perceived by the RGB-D camera during its motion; as more points of the OOI, previously occluded, enter into the field of view, it is possible that some of them induce a non-zero control signal. This discontinuity can be remedied by a first order low-pass filter.
\end{itemize}

To synthesize the total proposed control signal $\vct{V}_{cu}$, we calculate the angular velocity $\vect{\omega}_{j,k}$, which is induced by the virtual repulsive velocity $\vct{u}_{j,k}$ around an axis passing from $\vct{p}_{s_j}$ and defined by the cross product of the directions of $\vct{u}_{j,k}$ and $\vct{p}_{s_j}$. This angular velocity is given by:
\begin{equation}\label{eq:omega_pivot}
    \vect{\omega}_{j,k}\triangleq 
    \begin{cases}
          \frac{\vct{S}(\hat{\vct{p}}_{j,k} - \vct{p}_{s_j})}{\|\hat{\vct{p}}_{j,k}-\vct{p}_{s_j}\|^2} \vct{u}_{j,k}, \; &\text{if} \; \Big( \frac{\bar{\vct{p}}_{cs_j}^\intercal}{\|\bar{\vct{p}}_{cs_j}\|} \Big) \bar{\vct{p}}_{co_k} \in(0, \|\bar{\vct{p}}_{cs_j}\|) \\
          \vct{0}_{3 \times 1}, \; &\text{otherwise}
    \end{cases}.
\end{equation} 
By summing the $\vect{\omega}_{j,k}$-s acting on the $j$-th pivot point, we get the total angular velocity for the $j$-th OOI point, which is given by:
\begin{equation}\label{eq:omega_i}
    \vect{\omega}_{s_j} \triangleq \sum_{k=1}^{n_O} \vect{\omega}_{j,k} \in \mathbb{R}^3.
\end{equation}
Lastly, to synthesize the total control signal $\vct{V}_{cu}$, the $\vect{\omega}_{s_j}$ are superimposed after calculating the corresponding linear velocity at the end-effector. This superimposition is given by:
\begin{equation}\label{eq:v_c2}
\vct{V}_{cu} 
     \triangleq k_c \sum_{j=1}^{n_S}
     \begin{bmatrix}
        \vct{S}(\bar{\vct{p}}_{cs_j}) \\
        \vct{I}_3 
    \end{bmatrix}
     \vect{\omega}_{s_j} \in \mathbb{R}^6, 
\end{equation}
where $k_c \in \mathbb{R}_{>0}$ is a positive, tunable gain.

Taking the time derivative of the artificial potential \eqref{eq:art_potential}, we find that: $\dot{V} = - \frac{\|\bar{\vct{p}}_{cs_j}\|-\|\hat{\vct{p}}_{j,k}-\vct{p}_c\|}{\|\vct{p}_{s_j}-\hat{\vct{p}}_{j,k}\|}\vct{u}_{j,k}^\intercal\vct{u}_{j,k}$, which is less or equal than $0$, given that $\|\hat{\vct{p}}_{j,k}-\vct{p}_c\|\leq\|\bar{\vct{p}}_{cs_j}\|$, which is true by construction. This means that $V$ is bounded and that $\hat{r}_{j,k}$ is increasing within the area of influence, due to the fact that $V(\hat{r}_{j,k})$ is a decreasing function of $\hat{r}_{j,k}$. Therefore, $\vct{p}_{s_j}$ will not be occluded by the obstacle centered at $\vct{p}_{o,k}$. Furthermore, given that the linear velocity  of $\vct{V}_{cu}$ is orthogonal to $\bar{\vct{p}}_{cs_j}$, which implies that $\|\bar{\vct{p}}_{cs_j}\|$ remains constant, the maximum radius of visibility around $\vct{p}_{s_j}$ is increased with the proposed control signal. As a result, the progressive unveiling of more points of the OOI occurs in a chain-reaction manner, i.e. by unveiling progressively more and more OOI points.  

\subsection{Grasping Arm Control - Cutting Affordance Task} 
\label{subsec:grasp_contr}

To realise the grasping arm's control objective (Subsection \ref{subsec:grasp_prob}), reference velocity $\vct{V}_g$ is designed as a force/position controller, presented below:
 \begin{align} 
        \vct{v}_p &= - k_{pg} ( \vct{I}_{3 \times 3} - \vct{n}_c \vct{n}_c^\intercal ) \vct{e}_{pg} \ \in \mathbb{R}^3 \label{eq:gr_pos_contr} \\
        \vct{v}_f &= - \vct{n}_c \bigg( k_{fp} \vct{e}_{fg} + k_{fi} \int_{0}^{t} \vct{e}_{fg} \bigg)\ \in \mathbb{R}^3 \label{eq:gr_forc_contr},
\end{align}
where $\vct{e}_{pg} = \vct{p}_{g} - \vct{p}_{gd} \in \mathbb{R}^3$, $\vct{e}_{fg} = \vct{n}_c^\intercal f - f_d \in \mathbb{R}$ are the position and force errors, $k_{pg}, k_{fp}, k_{fi} \in \mathbb{R}$ are positive, control gains and 
\begin{equation} \label{eq:contr_dir}
        \vct{n}_c = \frac{ \vct{p}_{gd} - \vct{p}_{sb} }{ \| \vct{p}_{gd} - \vct{p}_{sb} \| }.
\end{equation}

\begin{figure}
    \centering
    \includegraphics[scale = .5]{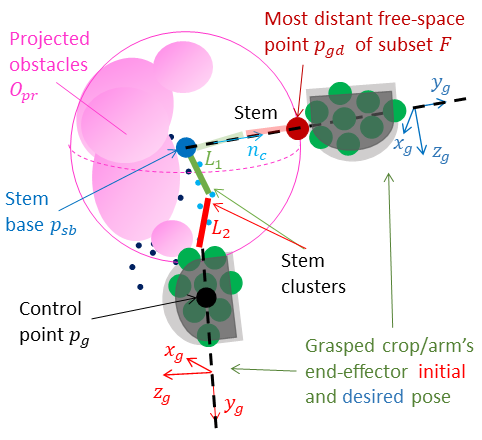}
    \caption{Grasping arm force/position and orientation control. Force control \eqref{eq:gr_forc_contr} applied at $\vct{n}_c$. Position control \eqref{eq:gr_pos_contr} applied at subspace $\vct{I}_{3 \times 3} - \vct{n}_c \vct{n}_c^\intercal$. Orientation control aligns $\vct{n}_c$ with $\vct{y}_g$.}
    \label{fig:grasp_contr}
\end{figure}

The force control signal \eqref{eq:gr_forc_contr} is applied in the direction of $\vct{n}_c$ \eqref{eq:contr_dir} to maximize the stem's stretching at the end of its motion (Fig. \ref{fig:grasp_contr})  and the position control signal \eqref{eq:gr_pos_contr} along the space orthogonal to $\vct{n}_c$.  Both signals contribute to the motion of 
the grasped crop towards its final position which is determined by the  projection of the space target $\vct{p}_{gd}$ on the space orthogonal to $\vct{n}_c$, while the position on the $\vct{n}_c$ direction is indirectly determined by the achievement of the stretching force.  

Regarding the orientation of the grasped crop, let the grasped crop's direction coincide with one of the axes of the grasping arm end-effector's frame $\{G\}$, which is assumed to be the y-axis $\vct{y}_{g} \in \mathbb{R}^3$ (Fig. \ref{fig:grasp_contr}). The following proposed signal: 
\begin{equation} \label{eq:rot_contr}
        \vct{v}_{g\omega} = - k_{og} ( \vct{n}_c \times \vct{y}_{g} ) \in \mathbb{R}^3
\end{equation}
with $k_{og}$ a positive gain, aligns the arm's end-effector/grasped crop's direction $\vct{y}_{g}$ with the stretched stem's $\vct{n}_c$, facilitating further the harvesting process by avoiding any stem obstructions, as well as introducing a more human-like manipulation of the target crop (see Fig. \ref{fig:human_harv}, where in the final configuration just before the stem's cutting, the human hand along with the grasped crop are aligned with the crop's stem).

Combining \eqref{eq:gr_pos_contr}, \eqref{eq:gr_forc_contr} and \eqref{eq:rot_contr}, the reference velocity $\vct{V}_g$ regarding the grasping arm's end-effector is:
\begin{equation} \label{eq:superimp_g}
    \vct{V}_g = \begin{bmatrix} \vct{v}_{gt} \\ \vct{v}_{g\omega} \end{bmatrix} = \begin{bmatrix} \vct{v}_p + \vct{v}_f \\ \vct{v}_{g\omega} \end{bmatrix}
\end{equation}
with $\vct{v}_{gt} = \vct{v}_p + \vct{v}_f \in \mathbb{R}^3$ the grasping arm's translational velocity.

\subsection{Bimanual robot motion scheduling and coordination} 
\label{subsec:biman_coord}

As the camera arm is initially positioned away from the region of reaching we initiate the reaching/unveiling camera arm motion until some preset thresholds before both arms are provided by their respective reference velocities $\vct{V}_g$ \eqref{eq:superimp_g} and $\vct{V}_c$ \eqref{eq:superimposition} to complete the task. Thresholds regard the arm's velocity and the camera's distance from the surface of the region's sphere with radius $r$. This ensures that a locally optimal view of OOI (i.e. stem) is achieved, and a more accurate ROI center $\vct{p}_{sb}$  and free space target $\vct{p}_{gd}$ can be calculated by the current point-cloud  to enable a successful  fulfillment of the bimanual task.  An enhancement of the bimanual motion is also proposed by superimposing the grasping arm's translational velocity to the camera arm reference velocity to assist in avoiding stem occlusions by the gripper motion.  

In particular, the translational part of the grasping arm's velocity $\vct{v}_{gt}$ is superimposed at the camera end-effector's velocity $\vct{V}_c$, while an angular velocity component is additionally applied to avoid $\vct{v}_{gt}$ affecting the stem centering. Therefore, the updated camera arm's velocity is given by:
    \begin{equation}\label{eq:final_vc}
        \vct{V}_{c}' = \vct{V}_{c} + \begin{bmatrix} \vct{v}_{gt} \\  \vct{S}(\vct{p}_c - \vct{p}_{sb})\vct{v}_{gt} \end{bmatrix}
    \end{equation}
Note that the angular part of $\vct{V}_g$ is not applied at the camera arm as it might induce velocities that do not correspond to the stem's velocity.


\section{Experimental Evaluation} \label{sec:exper}

\begin{figure}
    \centering
    \includegraphics[scale = .11]{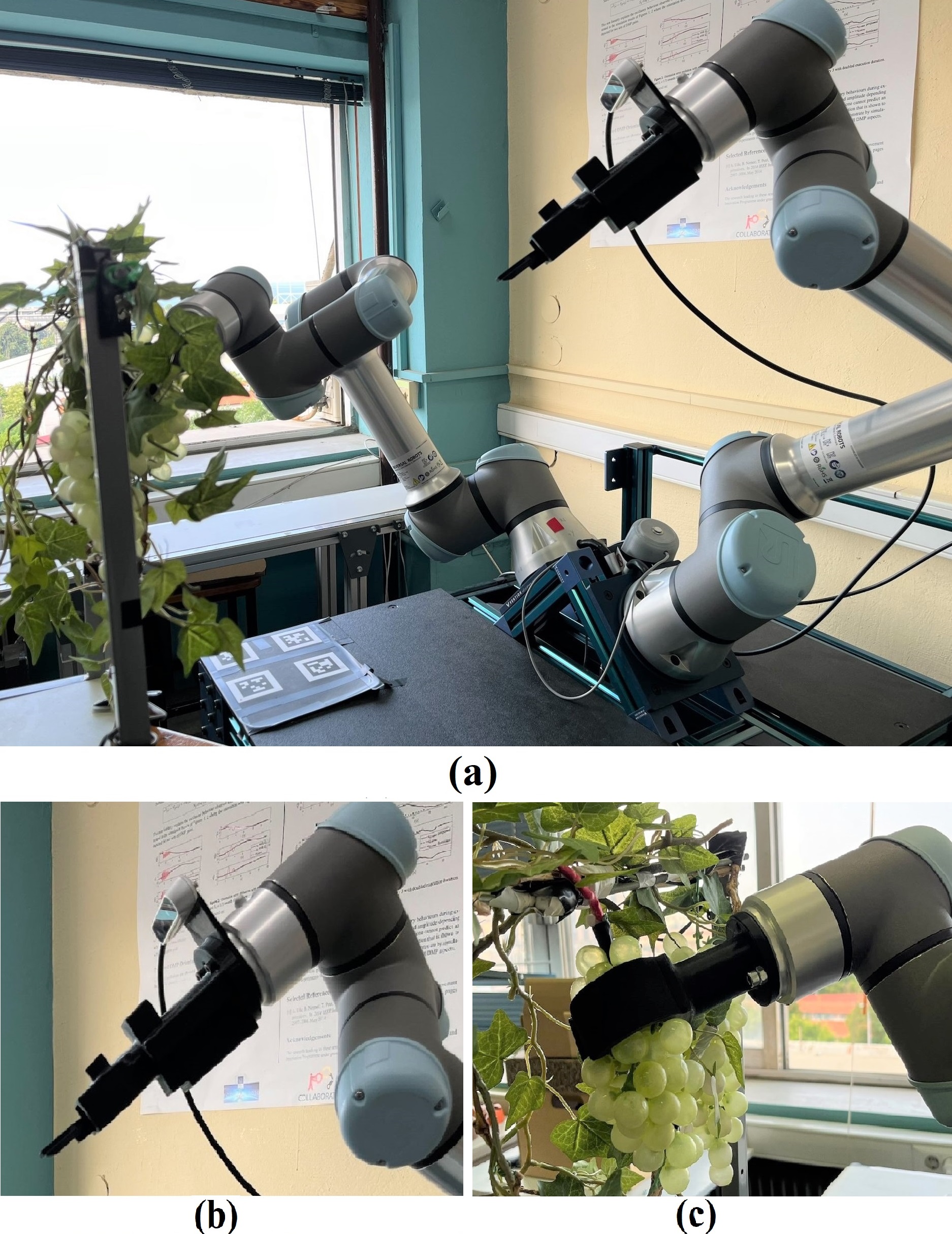}
    \caption{(a) Lab setup. (b) Camera arm. (c) Grasping arm initial state.}
    \label{fig:lab_setup}
\end{figure}

For the experimental evaluation of the proposed method, a lab setup was built with a mock-up vine illustrated in Fig. \ref{fig:lab_setup}, involving plastic vine branches and leaves as the surrounding obstacles and a plastic grape cluster as the target crop, with a stem of actual length $l_{act} = 0.07$ m. Identifying the point-cloud belonging to the stem is beyond the scope of this work. Thus for the experiments and in order to have this point-cloud available as the proposed method assumes we have used a red colored stem. Two UR5e robotic arms with a control cycle of $2$ ms were utilized. A 3d printed end-effector was used in the grasping arm to hold stably the mock up grape (Fig. \ref{fig:lab_setup}-(c)). On the camera arm (Fig. \ref{fig:lab_setup}-(b)), a Realsense D415 was mounted to capture the RGB image and the scene's point-cloud, with a frame rate of $30$ FPS and a resolution of $848 \times 480$ pixels and a 3d printed end-effector was attached at its tip to resemble a real cutting tool (Fig. \ref{fig:lab_setup}-(c)). To reduce computational complexity, the point-cloud was down-sampled by a factor of 3. 

The ROI is defined as a sphere with radius $r = 0.35$ m centered at the stem's base $\vct{p}_{sb}$, identified as in Subsection \eqref{subsec:scePC_proc}. The stem is reliably detectable only within a range of $0.7$ m from the camera. Any other points are obstacles, for which we use  spheres with radius $d_o = 0.001$ m. The control parameters were selected to be $k_{cp} = k_{co} = 1$ for \eqref{eq:control_signal_rr_world}, $d_a = 0.01$ m, $k_c = 0.00025$ for \eqref{eq:v_c2}. Regarding the thresholds that mark the beginning of the bimanual motion, we use the camera's distance from the ROI sphere a to be  $1.05 r$ and tip translational and angular velocities to be less than $0.01$ m/s and $0.025$ rad/s respectively. 
 
With target $\vct{p}_{gd}$ calculated from \eqref{eq:attr_point} and considering as obstacles all non-stem points residing in sphere $\text{S}(\vct{p}_{sb},r_o)$ with $r_o$ equal to the estimated  stem length $l = 0.0631$ m, the grasping arm's control parameters in \eqref{eq:superimp_g} are chosen to be $k_{pg} = 0.15$, $k_{og} = 0.3$, $k_{fp} = 0.001$, $k_{fi} = 0.0002$, with a desired force magnitude of $f_{d} = 3$ N. Notice that obstacle points are limited within $\text{S}(\vct{p}_{sb},r_o)$ only during the grasping arm's control, since in this case we are mostly interested in obstacles close enough to the stem, surrounding it; in the camera arm's case, we consider all non-stem points as obstacles to enable the stem's unveiling from any obstacles between itself and the camera. 

\begin{figure}[h!]
    \begin{subfigure}[t]{.49\linewidth }
        \centering
        \includegraphics[scale = .32]{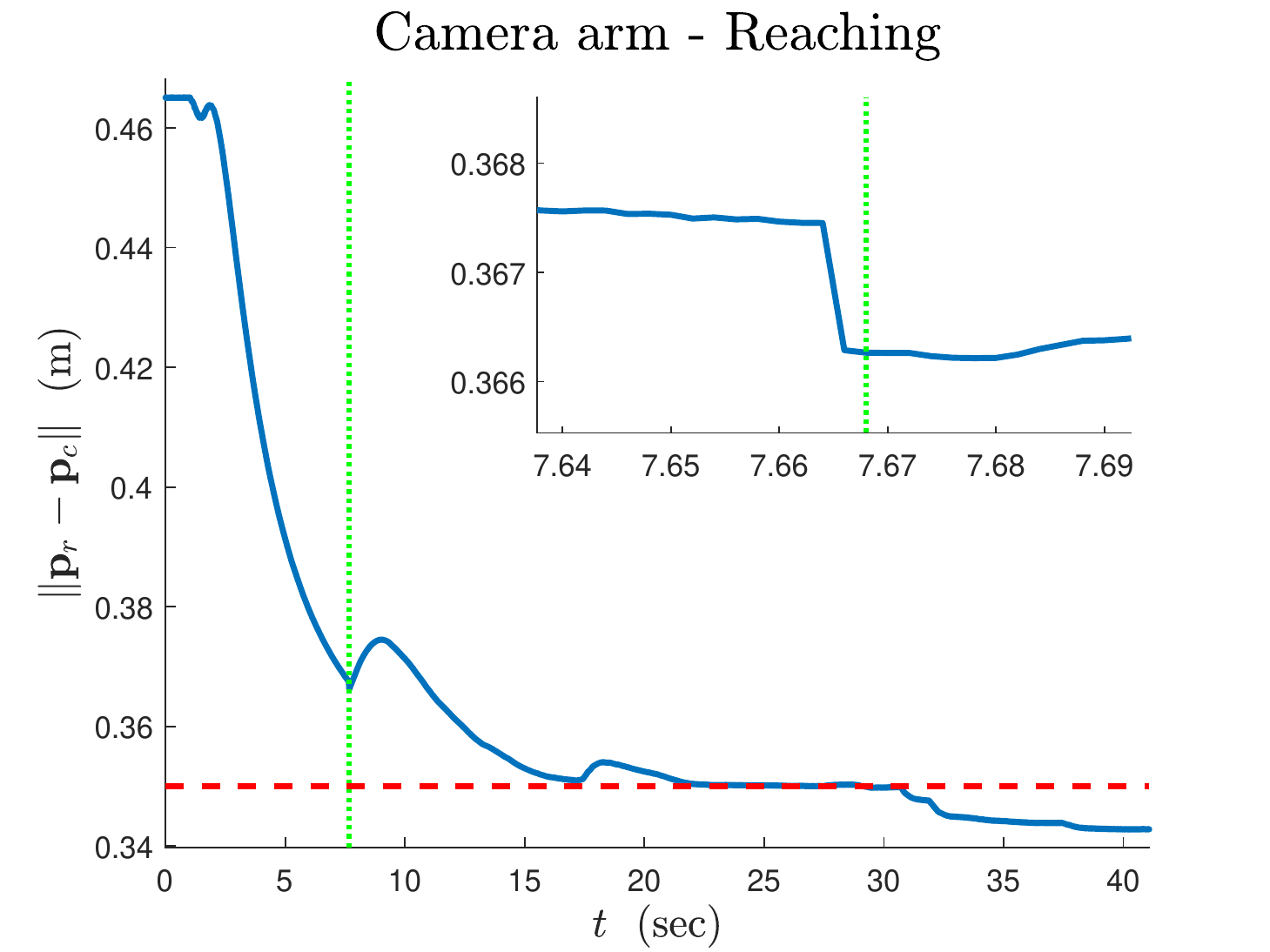}
        \caption{Reaching. Desired region radius $r = 0.35$ m - red dashed line.}
        \label{fig:exp_reach}
    \end{subfigure}
    \hfill 
    \begin{subfigure}[t]{.49\linewidth }
        \centering
        \includegraphics[scale = .32]{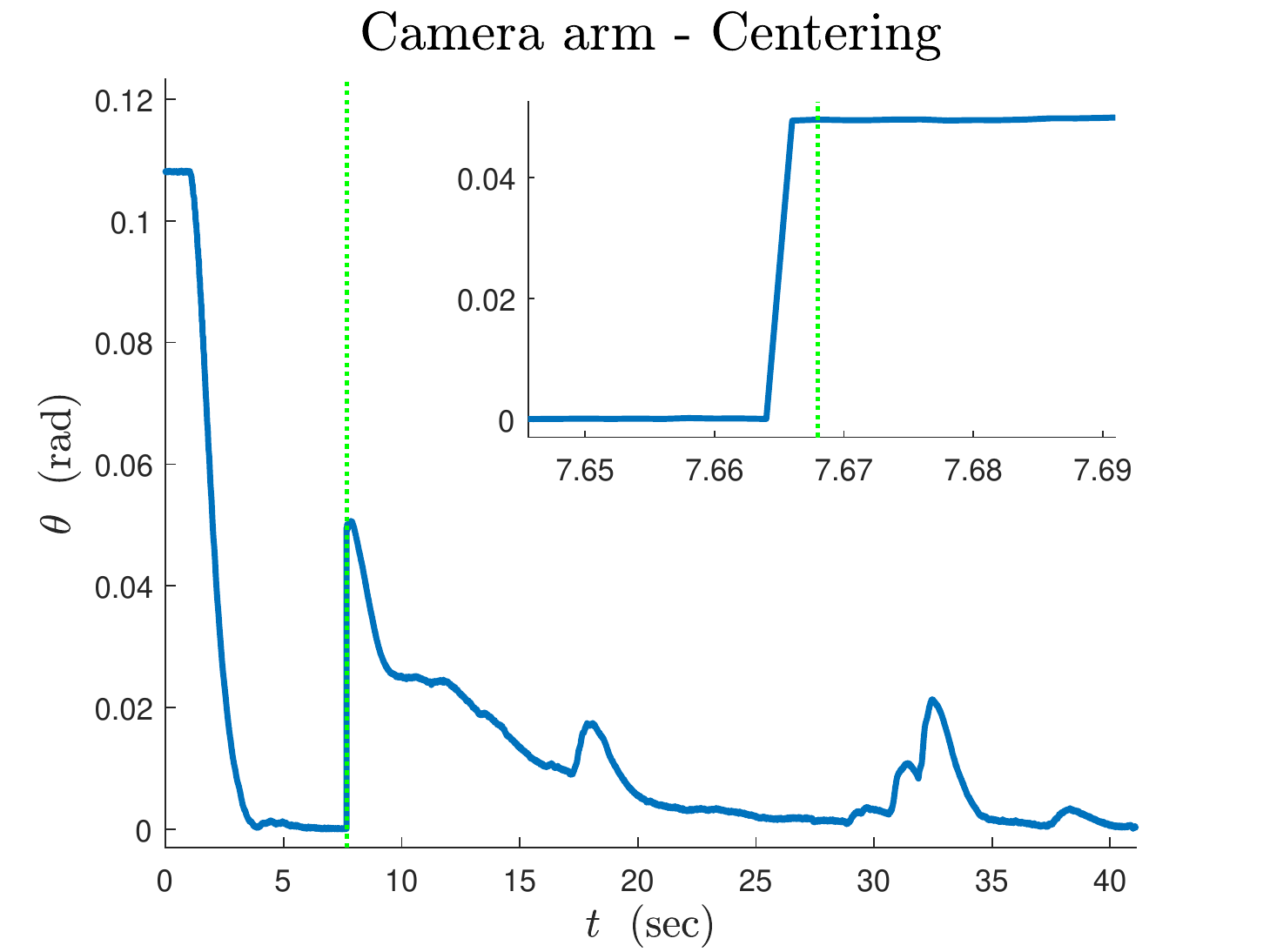}
        \caption{Centering. Angle $\theta$ between $\vct{p}_r - \vct{p}_c$ and camera’s view direction $\vct{z}_c$.}
        \label{fig:exp_center}
    \end{subfigure}
    \caption{Reaching with centering.}
    \label{fig:exp_reacent}
\end{figure}

\begin{figure}[h!]
    \centering
    \includegraphics[scale = .4]{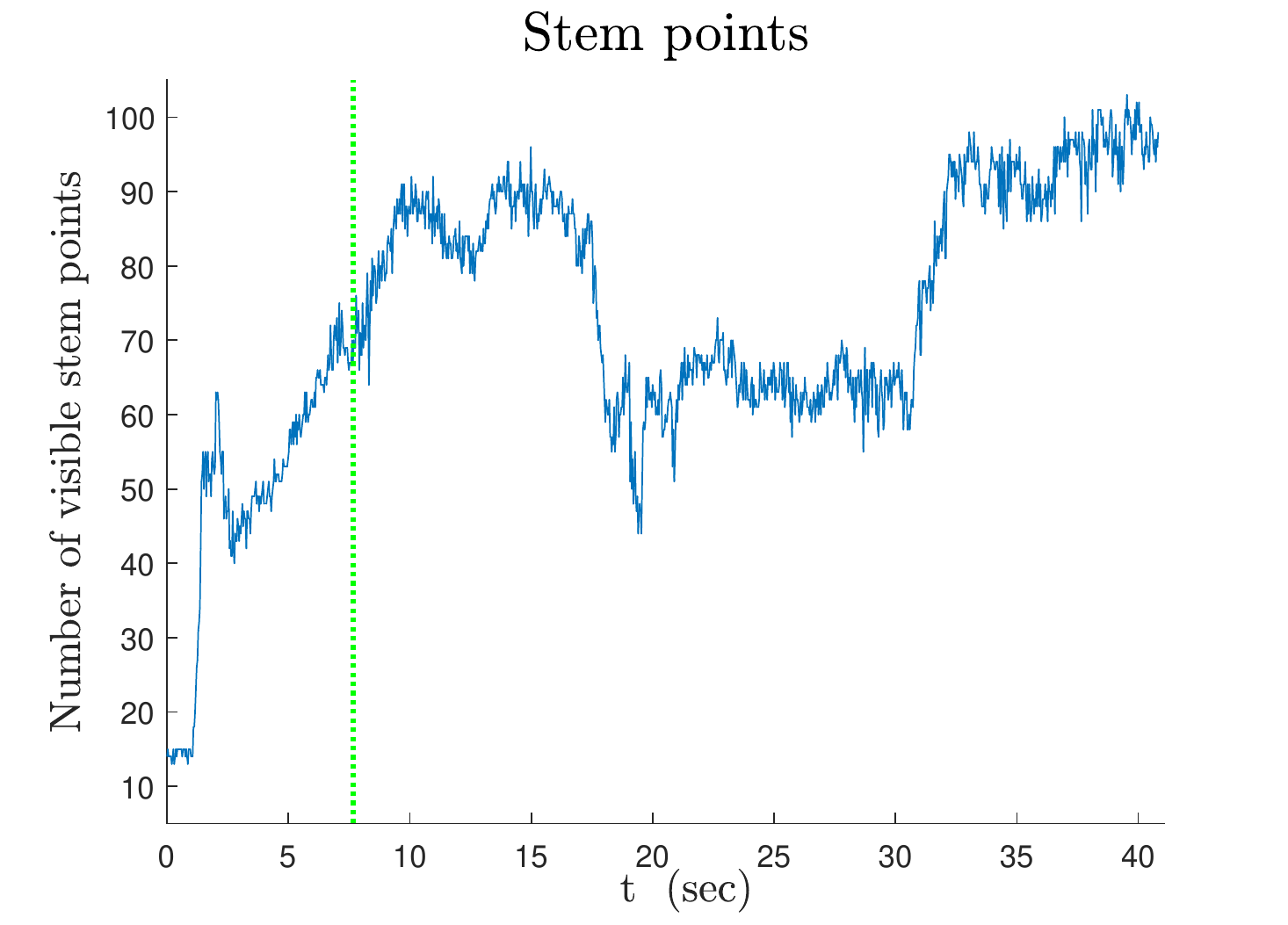}
    \caption{Unveiling. Visible stem points with respect to camera.}
    \label{fig:exp_unv}
\end{figure}

\begin{figure}[h!]
    \begin{subfigure}[t]{.49\linewidth }
        \centering
        \includegraphics[scale = .32]{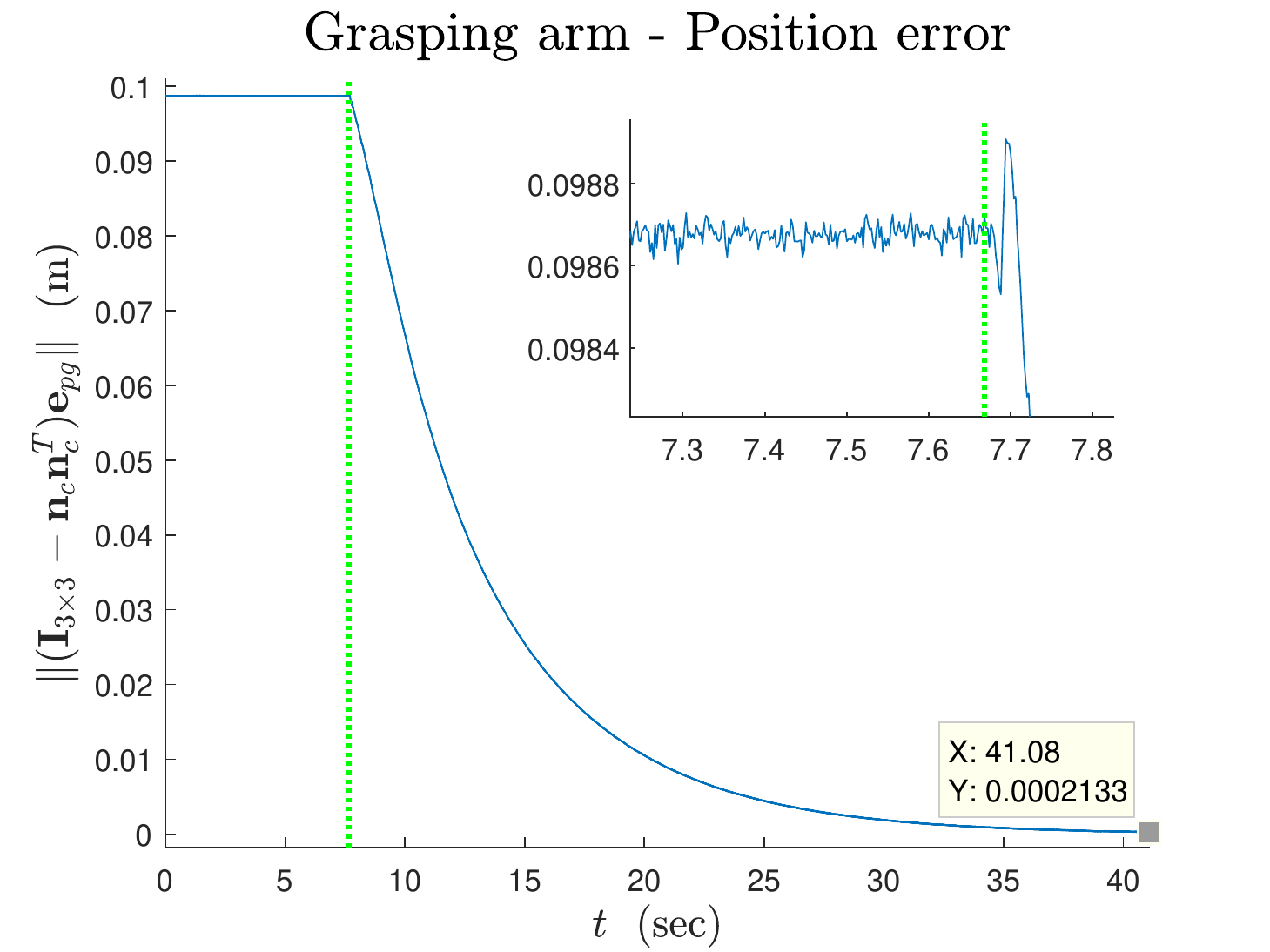}
        \caption{Position error.}
        \label{fig:exp_ep}
    \end{subfigure}
    \hfill 
    \begin{subfigure}[t]{.49\linewidth }
        \centering
        \includegraphics[scale = .32]{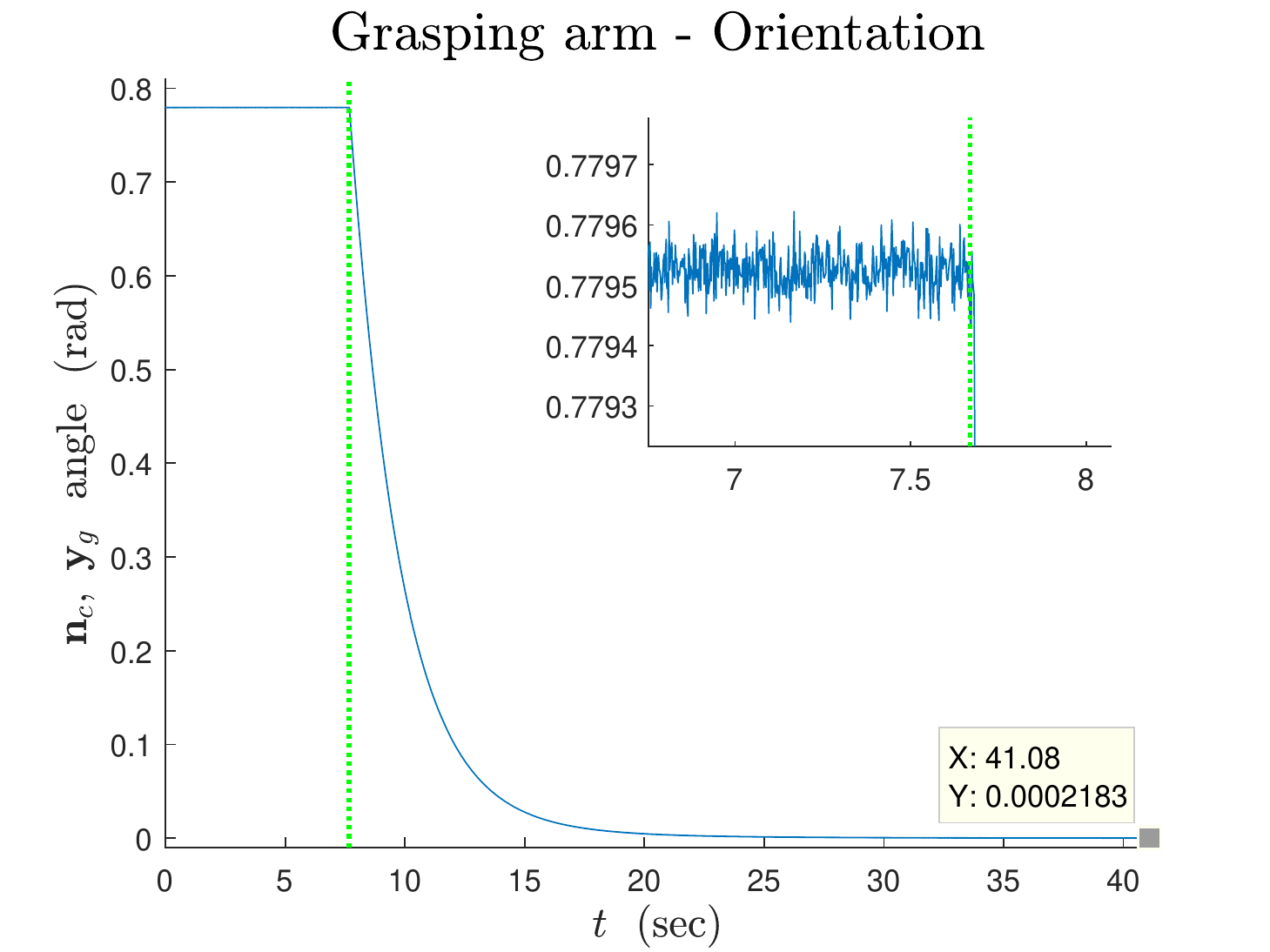}
        \caption{Grasped crop/Arm's end-effector alignment with stretched stem.}
        \label{fig:exp_eo}
    \end{subfigure}
    \caption{Grasping arm's position error and orientation alignment.}
    \label{fig:exp_epeo}
\end{figure}

\begin{figure}[h!]
    \centering
    \includegraphics[scale = .4]{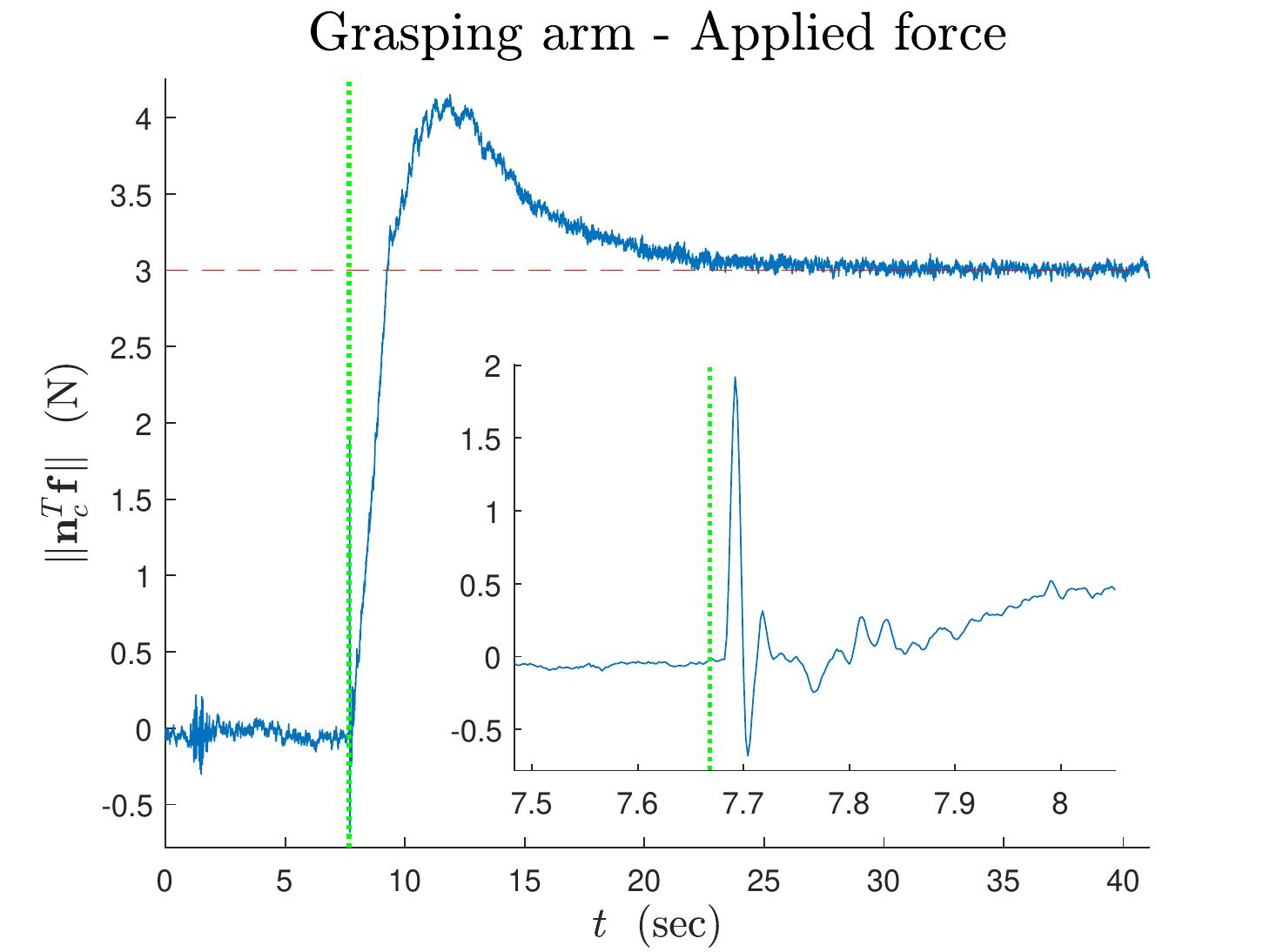}
    \caption{Applied force. Desired force magnitude $f_d$ - red dashed line.}
    \label{fig:exp_f}
\end{figure}

\begin{figure}[h!]
    \centering
    \includegraphics[scale = .6]{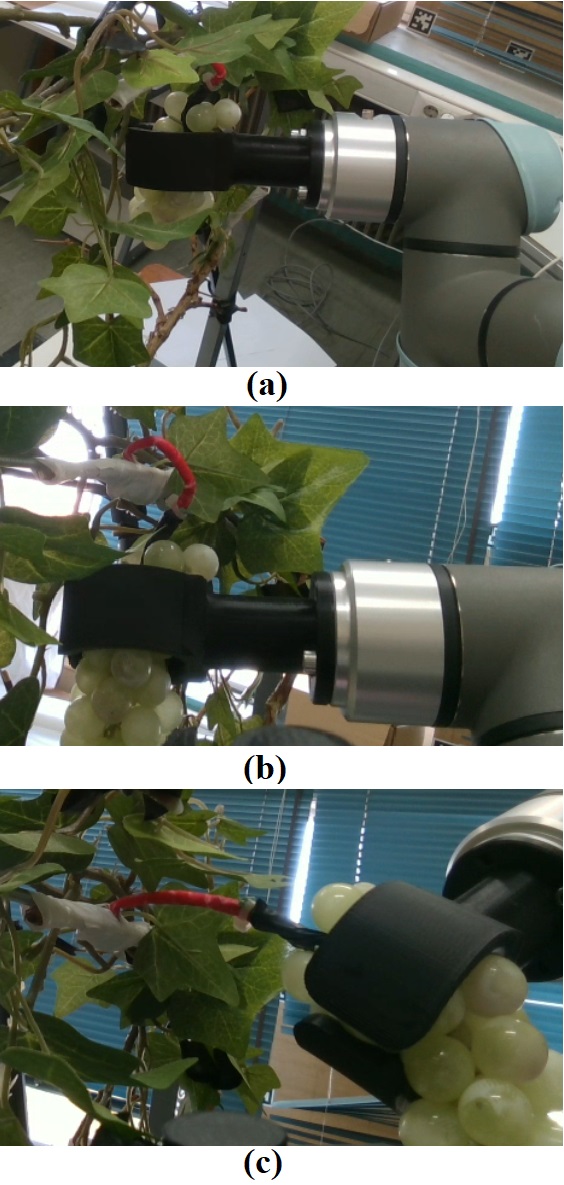}
    \caption{In-hand camera's viewpoint. (a) Process start. (b) Bimanual motion's start. (c) End of overall task.}
    \label{fig:exp_proc}
\end{figure}

Figures \ref{fig:exp_reacent} - \ref{fig:exp_proc} present the experimental results. In all figures, the green dotted line marks the start of the bimanual motion, that is at $7.67$ seconds. Figures \ref{fig:exp_reacent} - \ref{fig:exp_unv} depict the successful reaching of the desired region around the stem as well as its unveiling with respect to surrounding occlusions. The camera's distance from the region's sphere $\vct{p}_r - \vct{p}_c$ decreases and becomes even less than $r$ (Fig. \ref{fig:exp_reach}), that is the camera enters the ROI. Angle $\theta$ between $\vct{p}_r - \vct{p}_c$ and the camera’s view direction reaches a near-zero value (Fig. \ref{fig:exp_center}), thus the centering is achieved. Notice that just before the bimanual motion of the two arms starts (green dotted line), there is a discontinuity due to the recalculation of $\vct{p}_{sb}$, that is the center of ROI. Figure \ref{fig:exp_unv} shows the visible stem points, that increase steadily until the start of the bimanual motion as the camera moves for unveiling the stem. The drop in the number of visible points during the bimanual motion is due to the fact that the grasping arm moves inside the camera's field of view, thus hiding the stem momentarily. Nevertheless, this is corrected by the camera arm's control  continuing the stem's unveiling reaching an even higher number of unveiled points at the end. 


Figures \ref{fig:exp_epeo} - \ref{fig:exp_f} depict the grasping arm's related control objectives achievement. As seen in Fig. \ref{fig:exp_epeo}, the norm of the arm's end-effector position error projected in its corresponding action space $\vct{I}_{3 \times 3} - \vct{n}_c \vct{n}_c^\intercal$ as well as the angle between the arm’s end-effector/grasped crop’s direction $\vct{y}_g$ and the stretched stem’s $\vct{n}_c$ both become zero. Moreover, the stem is stretched successfully, since the desired force magnitude $f_d$ is reached (Fig. \ref{fig:exp_f}); notice that the spike at $7.67$ seconds (green dotted line) is caused by the transition from a static state to motion when the grasping arm's control is activated  \eqref{eq:gr_forc_contr}.

Figure \ref{fig:exp_proc} shows the camera's viewpoint at the start (Fig. \ref{fig:exp_proc}-(a)), at the start of the bimanual motion (Fig. \ref{fig:exp_proc}-(b)) and at the end of the task (Fig. \ref{fig:exp_proc}-(c)), enabling the stem cutting.
  

\section{Conclusions}

In this work, a bimanual control methodology for reaching a pre-cut state regarding the target crop's stem is presented. Control methods are designed for end-effector velocities for the camera and the grasping arm in order to reach and unveil the stem  as well as manipulate the crop to generate space around the stem. Lab experimentation with a mock-up vine setup and a plastic grape cluster validates the methodology's success. Future work involves the application of the proposed method in a vineyard. 


\bibliographystyle{IEEEtran}
\bibliography{main} 

\end{document}